\pdfoutput=1

\documentclass[11pt]{article}

\usepackage[]{EMNLP2023}

\usepackage{times}
\usepackage{latexsym}
\usepackage{booktabs}
\usepackage[T1]{fontenc}

\usepackage[utf8]{inputenc}

\usepackage{microtype}

\usepackage{inconsolata}

\title{Are Large Language Models Reliable Judges? \\ A Study on the Factuality Evaluation Capabilities of LLMs}

\author{Xue-Yong Fu, Md Tahmid Rahman Laskar, Cheng Chen, Shashi Bhushan TN \\
          Dialpad Canada Inc. \\
  \texttt{\{xue-yong,tahmid.rahman,cchen,sbhushan\}@dialpad.com}}

\begin{document}
\maketitle
\begin{abstract}
In recent years, Large Language Models (LLMs) have drawn significant attention due to their impressive emergent capabilities that were not observed in earlier language models. 
One emerging area where LLMs have been widely used in recent times is the utilization of LLMs as the evaluator of the texts generated by various generative models. 
In this paper, we also explore the possibility of whether LLMs are reliable in assessing the factual consistency of summaries generated by text generation models. We first propose a new approach to evaluate the factuality score 
using LLMs by utilizing one single LLM to perform all steps in the question-answering-based factuality scoring pipeline. Subsequently, we also study the performance of various LLMs to directly score the factuality. Our evaluation is conducted in traditional benchmarks by comparing their correlation with human annotations. Contrary to expectations, our findings reveal that none of the factuality metrics showed any significant correlations (e.g., coefficient scores greater than 0.3) to human evaluations of factuality for GPT-4 and PaLM-2, with the only exception being GPT-3.5 in two subcategories of factuality. 
Nonetheless, our findings are consistent across almost all factual error types, suggesting a fundamental limitation in the ability of current LLMs to assess factuality.
\end{abstract}
\section{Introduction}
Text summarization has significantly advanced through the utilization of pre-trained language models \cite{bert,liu2019text,lewis2020bart,t5,zhang2020pegasus,laskar2022domain}. However, a persistent concern with current models is their frequent inability to maintain factual consistency with the original documents they intend to summarize \cite{maynez2020faithfulness,fabbri2021summevalfaithfulness}. Consequently, establishing the factual accuracy of a summary continues to be the key for the evaluation of summarization models \cite{fabbri2021qafacteval,fabbri2022improving}. To resolve this issue, recent studies have utilized techniques like natural language inference, question-answering, or syntactic dependency as factuality evaluation metrics \cite{honovich2022true}. However, as highlighted by \citet{pagnoni2021understanding}, none of these automatic factuality metrics demonstrate a considerable correlation (i.e., fails to achieve a correlation score above 0.3) with human evaluations, pointing to the limited efficacy of these measures.

\begin{table*}[t!]
\small
    \centering
    \begin{tabular}{p{0.5\linewidth}p{0.45\linewidth}}
        \toprule
        \textbf{Prompt: QA-based Factuality Metric via LLMs} & \textbf{Prompt: LLM-based Factuality Scoring} \\
        \midrule
        \textbf{\# Answer Selection and Question Generation:} \newline
        \textit{From the following text, generate a question that can be answered within 1 or 2 words and also generate an answer that is either a noun phrase/named entity. \newline  %
            Text: Tom went to a baseball game tonight.   \newline 
            Output: \newline
            \{  \newline
            ``question": "When did Tom go to a baseball game?", \newline
            ``answer": "Tonight" \newline
            \}  \newline 
            Text: [SUMMARY]  \newline
            Output:  } 
\newline \newline 
               \textbf{\# Question Answering:} 
            
                  \textit{Answer the following question based on the given context. \newline %
           Question: [LLM Generated Question] \newline 
            Context: [ARTICLE]}
            &  \textit{Evaluate the quality of summaries written for a news article. Rate each summary on faithfulness. You should rate on a scale from 1 (worst) to 5 (best) without any explanation. \newline \newline
Article: Tom woke up at 7 AM and he went to school with his sister right away. \newline
        Summary: Tom went to school with his sister. \newline
         Faithfulness: 5 \newline \newline
        Article: [ARTICLE] \newline
        Summary: [SUMMARY] \newline
               faithfulness: }
     \\
        \bottomrule
    \end{tabular}
    \caption{Prompts for LLMs as  QA-based Factuality Evaluator and LLMs as Direct Faithfulness Scorer. In the QA-based factuality evaluator, the faithfulness score is measured based on the similarity between the initially selected answer (i.e., generated from the \textit{Answer Selection and Question Generation} step) and the final answer (i.e., the answer generated from the \textit{Question Answering} step)}
    \label{tab:prompts}
\end{table*}

The emergence and subsequent advancements of LLMs, such as  ChatGPT\footnote{\url{https://openai.com/blog/chatgpt}}, have transformed the landscape of natural language processing (NLP). ChatGPT-like LLMs \cite{palm2,touvron2023llama2,openai2023gpt4} have displayed impressive progress across a broad spectrum of NLP tasks, from text classification to generation, language translation, and beyond \cite{laskar2023systematic,laskar2023meeting}. Given the capabilities of these LLMs, our research explores the possibility of utilizing LLMs for the critical task of factual consistency evaluation \cite{dubois2023alpacafarm, liu2023gpteval, manakul2023selfcheckgpt, tang2022understanding, laban2023llms}.

To assess the factual consistency of a model, one common approach is the utilization of a question-answering (QA) pipeline \cite{huang2021factual}. Traditionally, the evaluation of factuality using QA systems has involved the use of separate, distinct models for each of the following tasks: \textit{answer selection}, \textit{question generation}, and \textit{question answering} \cite{huang2021factual}. However, this approach involves the intricate task of coordinating between these disparate models, potentially resulting in inefficiencies in real-world scenarios. Additionally, these models may fail to capture the comprehensive context necessary for optimal factuality evaluation. In response to these challenges, we propose a novel approach that substitutes the separate models with a singular and unified model using LLMs. 
In addition, we explore another approach where LLMs were directly asked to assess the factuality of a given summary. Meanwhile, we also address the potential risk of inaccurate high correlation measures~\cite{pagnoni2021understanding} by considering partial correlations, which are adept at controlling for confounding variables. In sum, this paper investigates the following Research Questions (RQ): 

\textbf{RQ 1:} Can the QA-based factuality metric be improved by utilizing LLMs? 

\textbf{RQ 2:} Can LLMs directly generate reliable faithfulness scores? 

\section{Related Work}
While neural abstractive summarization models can produce fluent summaries, they often generate factual inconsistencies \cite{honovich2022true}. In the early years of factual consistency evaluation, various unsupervised and weakly-supervised metrics have been used, which include relational triple-based, textual-entailment-based, as well as QA-based techniques \cite{huang2021factual}. Although the QA-based approach is a widely used technique for factuality evaluation, it requires separate models to perform different steps, such as question generation, answer selection, and finally, question answering. This makes the QA-based approach quite complicated and inefficient. In this regard, we study whether only one distinct LLM can be used to perform all steps in the QA-based factuality metric pipeline. Consequently, we also study whether LLMs can be directly used to predict the faithfulness score of the generated summary for a given article. 

Meanwhile, one major limitation in factuality evaluation is the lack of common benchmarks. This makes the comparison of various factuality metrics quite difficult. To address this issue, various benchmarks have been introduced recently for factual consistency evaluation, such as SummEval \cite{fabbri2021summevalfaithfulness} and FRANK \cite{pagnoni2021understanding}. These benchmarks are designed to evaluate various metrics on their ability to capture factual errors in abstractive summarization.  Among the available benchmarks, the FRANK benchmark is the largest one consisting of human-annotated factuality scores of summaries from diverse datasets. More specifically, it is a compilation of two datasets, CNN-DM \cite{nallapati2016abstractivecnndm} and XSUM \cite{narayan2018donxsum}, amalgamating outputs from nine distinct models across these datasets (5 models for CNN-DM and 4 models for XSUM). In total, the dataset comprises 2250 human-annotated judgments on different types of factual errors of model outputs. In addition, this benchmark addresses the false measurement of high correlations in various factuality metrics by introducing the partial correlation coefficients. 

In this paper, we also utilize the FRANK benchmark to evaluate the factual consistency of model-generated summaries by leveraging LLMs as the evaluator. Our paper diverges from that of~\citet{gao2023humanlike} in several key aspects. Notably, our research employs the FRANK dataset, encompassing the CNN-DM and XSUM datasets. In contrast,~\citet{gao2023humanlike} base their findings on the SummEval and Newsroom datasets. Additionally, our study presents results using partial correlation as opposed to the straightforward correlation employed by~\citet{gao2023humanlike}. This metric is adept at controlling for confounding variables, potentially mitigating the risk of inaccurate high correlation measures~\cite{pagnoni2021understanding}.


\section{Methodology}

In this section, we present our methods: (i) Using LLMs as QA-based factuality metric, and (ii) Using LLMs for direct factuality scoring. 
Below, we first present these methods. 

\textbf{(i) QA-based Factuality Metric via LLMs:} The reason we chose to incorporate LLMs into the QA-based factuality metric is that it is more reliable than most other existing automatic factuality metrics for assessing the factual consistency of a model \cite{huang2021factual}. 
The typical process of using QA-based systems as factuality evaluators is comprised of 3 tasks:

\textbf{(i) Answer Selection:} The commencement of this procedure involves extracting key points, referred to as ``answers'' from the provided summary. 
    
 \textbf{(ii) Question Generation:} After identifying the answers, the next step is to formulate questions based on these answers, using the summary as the context.

 \textbf{(iii) Question Answering:} The final step is responding to the generated questions using the input document as a reference. 

In this paper, contrary to the traditional approach of utilizing separate models to perform each task that makes the QA-based factuality evaluation process very complicated, we propose one single LLM to be used as the QA-based factuality metric evaluator to perform all steps. 
For prompt construction, we first evaluate various prompts in some samples and then select the one for our experiment that performs the best. We show our selected prompt for this task that we use in our experiments in Table \ref{tab:prompts}. 

In our prompt, we leverage the in-context learning principle and provide an associated example with our prompt to the LLMs to perform the first two tasks: initial answer selection and question generation. Since both the initial answer and the questions are required to be generated from the given summary (making both the question and the answer to have some dependencies between them), we unify these two steps together by asking the LLM to generate both the answer and the question simultaneously from the given summary. This makes the first two steps of the QA-based pipeline to be more efficient. Afterward, the generated question and the article are given as input to the LLM to generate the final answer. The evaluation process of the QA-based factuality metric depends on finding the similarity between the initially selected answer and the final answer. The higher the similarity, the more faithful the summary is being considered.

\textbf{(ii) Direct Faithfulness Scoring via LLMs:}  Similar to how we constructed prompts for the QA-based factuality metric evaluation, we first evaluate various prompts in a set of samples and select the one for full experiments that performs the best. With in-context example demonstrations, we prompt the target LLM to assess a provided summary based on faithfulness on a scale from 1 to 5 (our prompt is shown in Table \ref{tab:prompts}). 



\begin{table*}
\tiny
\setlength{\tabcolsep}{4pt} 
\centering
  \begin{tabular}{l|ccc|ccc|ccc|ccc}
\toprule
& \multicolumn{3}{c}{\textbf{Pearson $\rho$}} & \multicolumn{3}{c}{\textbf{Pearson p-value}} & \multicolumn{3}{c}{\textbf{Spearman $r$}} & \multicolumn{3}{c}{\textbf{Spearman p-value}}\\
\cmidrule(lr){2-4} \cmidrule(lr){5-7} \cmidrule(lr){8-10} \cmidrule(lr){11-13}
\textbf{Metric} & \textbf{PaLM-2} & \textbf{GPT-3.5} & \textbf{GPT-4} & \textbf{PaLM-2} & \textbf{GPT-3.5} & \textbf{GPT-4} & \textbf{PaLM-2} & \textbf{GPT-3.5} & \textbf{GPT-4} & \textbf{PaLM-2} & \textbf{GPT-3.5} & \textbf{GPT-4} \\
\midrule
 \textbf{Factuality Errors} & -0.0409 & -0.0016 & -0.0014 & 0.1050 & 0.9498 & 0.9561 & -0.0632 & -0.0259 & 0.0084 & 0.0121 & 0.3037 & 0.7390\\
 \midrule
 \textbf{Semantic  Frame  Errors} & -0.0416 & -0.0533 & -0.0386 & 0.0985 & 0.0343 & 0.1260 & -0.0005 & -0.0752 & -0.0494 & 0.9845 & 0.0028 & 0.0501\\ 
  \textbf{PredE} & -0.0057 & -0.0145 & -0.0044 & 0.8220 & 0.5650 & 0.8622 & 0.0928 & -0.0434 & -0.0290 & 0.0002 & 0.0848 & 0.2497\\
 \textbf{EntE} & -0.0211 & -0.0044 & -0.0212 & 0.4027 & 0.8617 & 0.4006 & 0.0645 & -0.0401 & -0.0327 & 0.0105 & 0.1117 & 0.1941\\
 \textbf{CircE} & -0.0307 & -0.0496 & -0.0444 & 0.2240 & 0.0491 & 0.0782 & 0.1044 & -0.0915 & -0.0419 & 0.0000 & 0.0003 & 0.0961\\ \midrule
 \textbf{Discourse  Errors} & -0.0177 & -0.0184 & -0.0185 & 0.4820 & 0.4649 & 0.4633 & -0.1073 & 0.0289 & 0.0065 & 0.0000 & 0.2522 & 0.7962\\
 
  \textbf{CorefE} & -0.0174 & -0.0222 & -0.0158 & 0.4897 & 0.3790 & 0.5306 & -0.0857 & 0.0158 & 0.0136 & 0.0007 & 0.5314 & 0.5890\\
 \textbf{LinkE} & -0.0057 & 0.0019 & -0.0173 & 0.8210 & 0.9385 & 0.4938 & 0.1424 & -0.0640 & -0.0567 & 0.0000 & 0.0110 & 0.0245\\ \midrule
 \textbf{Content  Verifiability  Errors} & 0.0185 & 0.0692 & 0.0335 & 0.4621 & 0.0060 & 0.1844 & 0.0011 & 0.0846 & 0.0359 & 0.9647 & 0.0008 & 0.1545\\
 \textbf{OutE} & 0.0302 & 0.0570 & 0.0472 & 0.2314 & 0.0237 & 0.0610 & 0.0212 & 0.0375 & 0.0300 & 0.3999 & 0.1373 & 0.2347\\
 \textbf{GramE} & -0.0187 & 0.0128 & -0.0297 & 0.4590 & 0.6130 & 0.2395 & 0.1103 & -0.0641 & -0.0397 & 0.0000 & 0.0110 & 0.1157\\


\bottomrule
\end{tabular}
\caption{Correlation scores for different LLMs as QA-based Factuality Metric Evaluator.}\label{tab:qa_correlation} 
\end{table*}

\begin{table*}
\tiny
\setlength{\tabcolsep}{4pt} 
    \centering
    \begin{tabular}{l|ccc|ccc|ccc|ccc}
        \toprule
       & \multicolumn{3}{c}{\textbf{Pearson $\rho$}} & \multicolumn{3}{c}{\textbf{Pearson p-value}} & \multicolumn{3}{c}{\textbf{Spearman $r$}} & \multicolumn{3}{c}{\textbf{Spearman p-value}}\\
\cmidrule(lr){2-4} \cmidrule(lr){5-7} \cmidrule(lr){8-10} \cmidrule(lr){11-13}
\textbf{Metric} & \textbf{PaLM-2} & \textbf{GPT-3.5} & \textbf{GPT-4} & \textbf{PaLM-2} & \textbf{GPT-3.5} & \textbf{GPT-4} & \textbf{PaLM-2} & \textbf{GPT-3.5} & \textbf{GPT-4} & \textbf{PaLM-2} & \textbf{GPT-3.5} & \textbf{GPT-4} \\
        \midrule
        \textbf{Factuality Errors} & -0.0898 & 0.0246 & 0.0915 & 0.0004 & 0.3302 & 0.0003 & -0.0921 & -0.0073 & 0.0579 & 0.0003 & 0.7737 & 0.0217 \\ \midrule
         \textbf{Semantic Frame Errors} & -0.0787 & 0.0111 & 0.0206 & 0.0018 & 0.6590 & 0.4139 & -0.0826 & 0.0980 & 0.0118 & 0.0010 & 0.0001 & 0.6384 \\ 
             \textbf{PredE} & -0.0465 & 0.0172 & -0.0266 & 0.0651 & 0.4945 & 0.2917 & -0.0108 & 0.3337 & -0.0265 & 0.6687 & 0.0000 & 0.2934 \\ 
         \textbf{EntE} & -0.0641 & 0.0113 & -0.0177 & 0.0109 & 0.6554 & 0.4817 & -0.0569 & 0.1801 & -0.0243 & 0.0240 & 0.0000 & 0.3356 \\ 
         \textbf{CircE} & -0.0663 & 0.0266 & 0.0004 & 0.0084 & 0.2909 & 0.9884 & -0.0503 & 0.3702 & -0.0246 & 0.0459 & 0.0000 & 0.3288 \\ \midrule
        \textbf{Discourse Errors} & -0.0641 & 0.0178 & -0.0376 & 0.0110 & 0.4806 & 0.1355 & -0.0484 & -0.2273 & -0.0332 & 0.0546 & 0.0000 & 0.1879 \\ 
                  \textbf{CorefE} & -0.0632 & 0.0165 & -0.0345 & 0.0121 & 0.5131 & 0.1712 & -0.0519 & -0.2700 & -0.0215 & 0.0394 & 0.0000 & 0.3947 \\ 
         \textbf{LinkE} & -0.0520 & 0.0257 & -0.0440 & 0.0390 & 0.3086 & 0.0805 & -0.0219 & 0.2827 & -0.0499 & 0.3849 & 0.0000 & 0.0477 \\  \midrule
         \textbf{Content Verifiability Errors} & -0.0147 & 0.0316 & 0.0184 & 0.5612 & 0.2107 & 0.4662 & -0.0071 & 0.0148 & 0.0190 & 0.7784 & 0.5568 & 0.4510 \\ 
    
         \textbf{OutE} & -0.0131 & 0.0267 & 0.0468 & 0.6033 & 0.2891 & 0.0633 & -0.0052 & -0.0447 & 0.0483 & 0.8357 & 0.0761 & 0.0551 \\ 
         \textbf{GramE} & -0.0497 & 0.0285 & -0.0716 & 0.0488 & 0.2575 & 0.0045 & -0.0298 & 0.2893 & -0.0874 & 0.2377 & 0.0000 & 0.0005 \\ 
       

        \bottomrule
    \end{tabular}
    \caption{Correlation scores for different LLMs as Faithfulness Scorer.}
    \label{tab:faithful_correlation} 
\end{table*}

\section{Experiments}
In this section, we first present the LLMs that we study in this paper, followed by defining the evaluation metrics and finally the experimental results. 

\subsection{Models}
We use the following LLMs for evaluation.  

\textbf{GPT-3.5:} GPT-3.5, also known as ChatGPT, is a transformer-based \cite{vaswani2017attention} auto-regressive model developed by OpenAI that was pre-trained on a vast amount of textual data via supervised learning alongside reinforcement learning with human feedback. We use the \textit{gpt3.5-turbo-0613} version of this model via OpenAI\footnote{\url{https://platform.openai.com/docs/models}}. 

\textbf{GPT-4:} GPT-4 \cite{openai2023gpt4} is the latest addition to the GPT series models by OpenAI that is touted as being more reliable, creative, and able to handle much more nuanced instructions than GPT-3.5. However, GPT-4 is about 25x more costly than GPT-3.5 while being significantly slower. We use the \textit{gpt4-0613} version of this model via OpenAI. 

\textbf{PaLM-2:} It is also a transformer-based language model proposed by Google that exhibits enhanced reasoning capabilities and improved computing efficiency. We use the \textit{text-bison@001} version of this model through Google's Vertex API\footnote{\url{https://cloud.google.com/vertex-ai/docs/generative-ai/model-reference/text}}. 


\subsection{Evaluation Metrics}
While previous studies, such as \citet{gao2023humanlike}, have indicated the potential of automatic metrics in assessing factuality, not accounting for confounding variables associated with system and dataset properties in some contexts might influence the perceived correlations \citet{pagnoni2021understanding}. In contrast, our experiment addresses this concern by incorporating partial correlation coefficients, leveraging the FRANK benchmark {\cite{pagnoni2021understanding}}. The FRANK benchmark not only contains data from diverse datasets but also features a comprehensive typology of factual errors, allowing for a more nuanced understanding of the inaccuracies in generated summaries. As given in the FRANK benchmark, we measure partial correlation in terms of the following: 

\begin{enumerate}
    \item \textbf{Factuality Errors:} This is the overall factuality error. 
    
   \item \textbf{Semantic Frame Errors:} Errors that occur due to the incorrect understanding of the relationships and roles in a situation or event. Example: \textit{Predicate Errors}, \textit{Entity  Errors} and \textit{Circumstance Errors}. 
 
    \begin{itemize}
    \item \textbf{Predicate Errors (PredE):}  Incorrect or misrepresented predicates in summaries. 

    \item \textbf{Entity Errors (EntE):} Wrong entities mentioned.

   \item \textbf{Circumstance Errors (CircE):}  Inaccurate details regarding the circumstances of an event.

    \end{itemize}

\item \textbf{Discourse Errors:} It refers to incorrect links between different parts of a summarized text. Example: \textit{Coreference Errors} and \textit{Discourse Link Errors}. 

        \begin{itemize}

\item \textbf{Coreference Errors (CorefE):} Refers to incorrect references (e.g., pronoun).

\item \textbf{Discourse Link Errors (LinkE):}  Errors in connecting statements logically within a discourse.

    \end{itemize}

\item \textbf{Content Verifiability Errors:} These errors arise when the summaries cannot be verified for accuracy due to a lack of supporting evidence. Example: \textit{Out of Article Errors} and \textit{Grammatical Errors}.

\begin{itemize}
    \item 
\textbf{Out of Article Errors (OutE):} Statements containing information not present in the referenced source. 

\item \textbf{Grammatical Errors (GramE):}  Grammatical mistakes that make sentences factually incorrect.

\end{itemize}

\end{enumerate}

\subsection{Results and Discussion}

For the QA-based factuality, the common metrics used to measure the correlation include the Exact Match and the word F1 scores. However, the Exact Match could be excessively stringent. Thus, we opt for the word F1 which offers a more balanced evaluation for answer overlap.

\paragraph{(i) LLM as QA-based Factuality Metrics:} We show our results for the QA-based factuality evaluation in Table \ref{tab:qa_correlation}. For overall factuality (referred to as ``Factuality Errors"), only PaLM-2 displays a statistically significant p-value of $0.0121$ for the Spearman partial correlation. This indicates that there is no linear correlation between human judgment and the LLM-QA score, as the correlation coefficient is $-0.0632$. For the majority of factuality error subcategories, PaLM-2, GPT-3.5 and GPT-4 do not have statistically significant p-values for the Pearson correlation coefficient. 
However, the correlation values for all are very close to zero, which indicates no linear correlation between human judgment and the LLM-QA score even for the subcategories. In terms of the Spearman correlation coefficient that is capable of detecting non-linear relationships, PaLM-2 exhibits a statistically significant but very weak correlation (greater than 0.1 but less than 0.3) with human judgment in Discourse Errors, CircE, GramE, and LinkE, where the absolute value exceeds 0.1.

\paragraph{(ii) LLM as Direct Faithfulness Scorer:} Table \ref{tab:faithful_correlation} shows the correlation coefficients calculated between the factuality scores assigned by LLMs and the scores corresponding to different types of human-annotated errors. 
In terms of error subcategories, we see PaLM-2 doesn't show any correlation with high p-values and close to zero coefficients. 
Both GPT-3.5 and GPT-4 also do not have any significant Pearson correlation scores. But interestingly GPT-3.5 shows statistically significant Spearman correlation scores for Discourse Errors ($-0.2273$), PredE ($0.3337$), EntE ($0.1801$), CircE ($0.3702$), GramE ($0.2893$), CorefE ($-0.27$) and LinkE ($0.2827$). The observed negative correlation is worrisome, as it could suggest inherent issues with the model's reliability as a faithfulness scorer.



\section{Conclusion}
The central objective of this research was to assess the effectiveness of various LLMs, specifically GPT-3.5, GPT-4, and PaLM-2 in the evaluation of factuality in text summarization tasks. In addition to directly using LLMs to evaluate the factuality of a summary, we also introduce a novel approach that utilizes one single LLM to perform various steps of the QA-based factuality scoring pipeline. Contrary to expectations, our findings revealed that none of the approaches showed a significant correlation (with a coefficient greater than 3) to human evaluations of factuality for most LLMs, with the only exception happening while directly generating the LLM faithfulness scores by GPT-3.5 in two subcategories of factuality: PredE and CircE. Nonetheless, the result is consistent across almost all factual error types, suggesting a fundamental limitation in the ability of current LLMs to effectively assess factuality.

While previous studies, such as \citet{gao2023humanlike}, indicated the potential of automatic metrics in assessing factuality, our findings suggest that it is essential to consider possible dataset biases \citet{pagnoni2021understanding}. In some contexts, not accounting for confounding variables associated with system and the dataset properties might influence the perceived correlations. To provide a more nuanced perspective, we recommend utilizing partial correlation coefficients to control for these variables. Our study calls for an exploration into the inherent deficiencies of current language models in maintaining factual consistency and sheds light on the necessity for developing more accurate and comprehensive models and methods for factuality evaluation. 

In the future, we will study the factuality evaluation capabilities of LLMs using other benchmarks \cite{laban2022summac,wang2023survey}, as well as on noisy datasets \cite{fu-etal-2022-entity,khasanova2022developing,laskar2022auto,laskar-etal-2022-blink,laskar-etal-2023-ai-coach-assist,manderscheid-lee-2023-predicting}, alongside investigating new approaches, such as the utilization of few-shot learning \cite{GPT3}, other prompting strategies \cite{liu2023pre}, and whether fine-tuning open-source LLMs \cite{touvron2023llama,touvron2023llama2,zhao2023survey} for factuality evaluation leads to a better factuality evaluator. 

\section*{Limitations}
The closed-source models that have been used in this paper are continuously updated. This may lead to the potential deprecation or unavailability of the older versions of the models with the release of newer versions. Thus, there might be some variations in the results while replicating our study. 

\bibliography{custom}
\bibliographystyle{acl_natbib}





\end{document}